\documentclass[10pt,twocolumn,letterpaper]{article}

\usepackage{cvpr}              
\usepackage{url}
\usepackage{multirow}
\usepackage{booktabs}
\usepackage{pifont}
\usepackage{xcolor}
\usepackage{graphicx}
\usepackage{amsmath}
\usepackage{algorithm}
\usepackage{mathtools}
\usepackage{algorithm}
\usepackage{algpseudocode}
\usepackage{listings}
\usepackage{wrapfig}
\usepackage{subcaption}
\usepackage{xcolor,colortbl}
\newcommand{\cmark}{\color{green}\ding{51}}%
\newcommand{\xmark}{\color{red}\ding{55}}%

\usepackage{amssymb}
\usepackage{bbm}

\newcommand{\Mat}{\boldsymbol}

\newcommand{\real}{\mathbb{R}}

\definecolor{top1}{RGB}{255,179,179}
\definecolor{top2}{RGB}{255,217,179}
\definecolor{top3}{RGB}{255,255,179}

\definecolor{cvprblue}{rgb}{0.21,0.49,0.74}
\usepackage[pagebackref,breaklinks,colorlinks,citecolor=cvprblue]{hyperref}

\def\confName{CVPR}
\def\confYear{2024}

\title{Pose-Free Generalizable Rendering Transformer}

\author{%
  Zhiwen Fan\textsuperscript{1*}, Panwang Pan\textsuperscript{2*}, Peihao Wang\textsuperscript{1}, Yifan Jiang\textsuperscript{1}, \\ Hanwen Jiang\textsuperscript{1},  Dejia Xu\textsuperscript{1}, Zehao Zhu\textsuperscript{1}, Dilin Wang\textsuperscript{3}, Zhangyang Wang\textsuperscript{1}\\
  {\textsuperscript{1}University of Texas at Austin, \textsuperscript{2}ByteDance \textsuperscript{3}Meta} \\
  \small{\texttt{\{zhiwenfan,atlaswang\}@utexas.edu}} \\
}

\begin{document}

\twocolumn[{%
\renewcommand\twocolumn[1][]{#1}%
\maketitle
\begin{center}\vspace{-7mm}
    \includegraphics[width=1\textwidth]{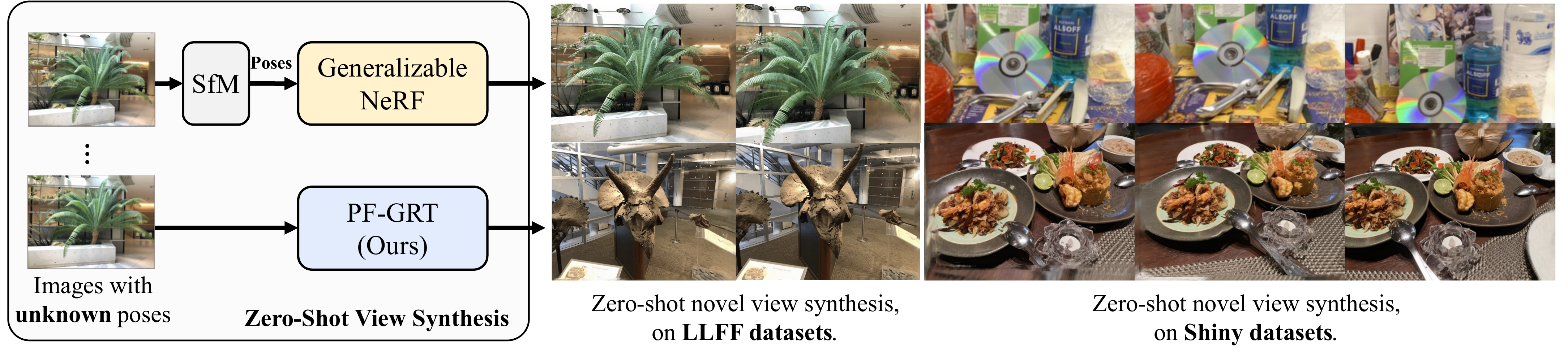}\vspace{-4mm}
    \captionof{figure}{Inference pipeline of our \underline{\textbf{P}}ose-\underline{\textbf{F}}ree \underline{\textbf{G}}eneralizable \underline{\textbf{R}}endering \underline{\textbf{T}}ransformer (\textbf{PF-GRT}), which facilitates novel view synthesis without the need for computing camera poses (left). We present zero-shot (generalizable) rendering results on various datasets (right).}
    \label{fig:teaser}
\end{center}
}]

\begin{abstract}
In the field of novel-view synthesis, the necessity of knowing camera poses (e.g., via Structure from Motion) before rendering has been a common practice. However, the consistent acquisition of accurate camera poses remains elusive, and errors in pose extraction can adversely impact the view synthesis process.
To address this challenge, we introduce \textbf{PF-GRT}, a new \underline{P}ose-\underline{F}ree framework for \underline{G}eneralizable \underline{R}endering \underline{T}ransformer, eliminating the need for pre-computed camera poses and instead leveraging feature-matching learned directly from data.
PF-GRT is parameterized using a local relative coordinate system, where one of the source images is set as the origin.
An OmniView Transformer is designed for fusing multi-view cues under the pose-free setting, where unposed-view fusion and origin-centric aggregation are performed.
The 3D point feature along target ray is sampled by projecting onto the selected origin plane.
The final pixel intensities are modulated and decoded using another Transformer.
PF-GRT demonstrates an impressive ability to generalize to new scenes that were not encountered during the training phase, without the need of pre-computing camera poses.
\renewcommand{\thefootnote}{\fnsymbol{footnote}}
\footnotetext[1]{Equal Contribution.}
\renewcommand{\thefootnote}{\arabic{footnote}}
Our experiments with \underline{zero-shot} rendering on the LLFF, RealEstate-10k, Shiny, and Blender datasets reveal that it produces superior quality in generating photo-realistic images. Moreover, it demonstrates robustness against noise in test camera poses. Code is available at \url{https://zhiwenfan.github.io/PF-GRT/}.
\end{abstract}

\section{Introduction}
\begin{figure*}[t]
\vspace{-1em}
\vskip 0.2in
\begin{center}
    \centering
    \includegraphics[width=0.99\linewidth]{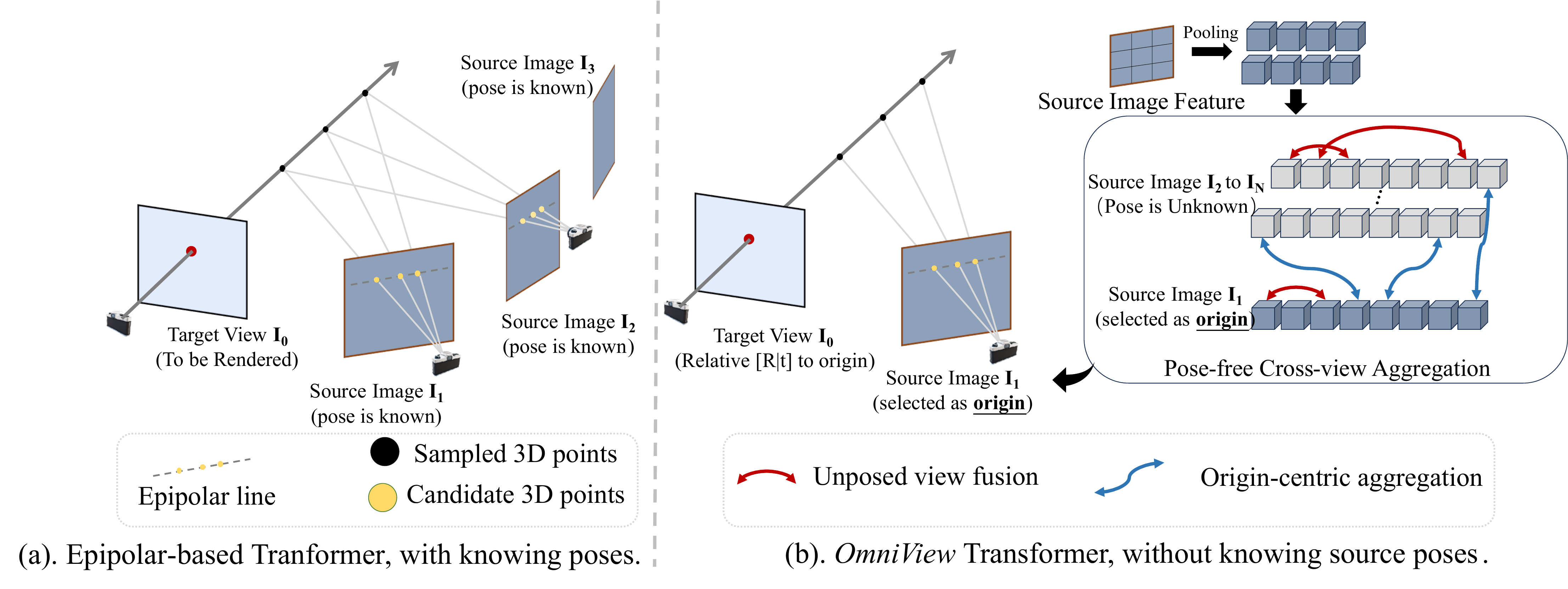}
\vspace{-0.5em}
\caption{\textbf{Epipolar Transformer vs. OmniView Transformer.} The Epipolar Transformer requires camera poses to search for candidate matching along the epipolar line. The OmniView Transformer finds the correspondences using global feature matching by using Unposed View Fusion and Origin-Centric Aggregation, which does not rely on camera poses.}
\label{fig:omniview}
\end{center}
\vspace{-1.5em}
\end{figure*}
Novel view synthesis, as demonstrated by recent works~\cite{yu2021pixelnerf, wang2021ibrnet, liu2022neural, suhail2022generalizable, wang2022attention}, has showcased the capability to generate new views on unseen scenes in a feed-forward manner. Despite their effectiveness, the prerequisite of camera poses for each view to offer explicit geometric priors between scene structures and 2D pixels is a common practice utilizing Structure from Motion (SfM)~\cite{schoenberger2016sfm} before rendering. However, accurate camera poses not only complicate the synthesis pipeline but also restrict applicability in scenarios where precise camera information is unavailable or difficult to obtain. In some cases, inaccurate pose estimation propagates its error to the renderer, adversely reducing the synthesized image quality. 
One could bypass the demand for camera poses by adopting only a single image to learn generalizable NeRFs (e.g., PixelNeRF~\cite{yu2021pixelnerf}), and render the target image from the constructed feature volume. 
On the other hand, Scene Representation Transformer (SRT)\cite{sajjadi2022scene} and RUST\cite{sajjadi2023rust} have pioneeringly explored the representation of multiple images as a ``set latent scene representation'' and generate novel views even in the presence of flawed camera poses or without any pose information. However, these works still face challenges: scene reconstruction under a single input is highly ill-posed and fails easily in in-the-wild scenes, while the latent representation results in blurred rendering outcomes with a lower resolution (e.g., 128$\times$128 in SRT), limiting their applicability in achieving photorealistic rendering.
In this work, we take a step forward, unifying global feature matching with Transformer~\cite{vaswani2017attention} and the image-based rendering (IBR)~\cite{hedman2018deep} for \textbf{photo-realistic novel view synthesis without the need for camera poses and per-scene optimization}, all accomplished in a feed-forward pass. 

Our proposed framework, PF-GRT, is parameterized by a local coordinate system, where one unposed view is used as a starting point (origin), and the target view to be rendered is defined by a relative transformation.
Another key design element comes from using Transformer for global feature matching and fusing the multi-view cues: the proposed \textit{OmniView Transformer} sequentially performs Unposed View Fusion and Origin-Centric Aggregation without the requirement of camera poses for computing epipolar constraints.
Pixel-aligned 3D feature along each target ray is sampled by projecting onto the origin plane of the local system.
The final pixel intensities along each target ray are modulated and decoded using another transformer, taking into account all warped 3D point features.

In \underline{training}, PF-GRT is optimized using a large-scale dataset~\cite{wang2021ibrnet} with calibrated multi-view images; the starting view (origin) is randomly selected, and the ground-truth poses are converted into relative for applying the photometric losses.
In \underline{inference}, our method selects any unposed source image as origin (the root of relative coordinate system), and can render any free-view by specifying a relative transformation relative to the origin. This is achieved without the pre-computed poses in an unseen scene.

Comprehensive experiments on diverse real-world scenes, including LLFF, RealEstate10K, Shiny, and object-level Blender datasets demonstrate our approach achieves superior zero-shot rendering quality (see Figure~\ref{fig:teaser}), surpassing prior pose-free generalizable NeRFs with a large-margin.
Additionally, our method demonstrates superior robustness against camera pose noise in new scenes.

Our major contributions are encapsulated as follows:
\begin{itemize}
\item We introduce a new formulation for generating novel views, by unifying the pixel-aligned feature of the target pixel with Image-based Rendering (IBR), eliminating the need for pose annotations among test scenes, and streamlining the rendering process.
\item We propose an efficient \textit{OmniView Transformer} to aggregate multi-view features with adaptive weights, by broadening the epipolar constraint to encompass all source pixels. A source-conditioned modulation layer is integrated to handle projection occlusions, and pixel intensities are decoded by another Transformer.
\item Trained with large-scale multi-view datasets, PF-GRT is able to render photo-realistic novel views in unseen scenes in a feed-forward pass. It significantly outperforms other pose-free, generalizable neural rendering techniques in diverse datasets.
\end{itemize}

\section{Related Works}

\begin{figure*}[t]
\vspace{-1em}
\vskip 0.2in
\begin{center}
    \centering
    \includegraphics[width=0.99\linewidth]{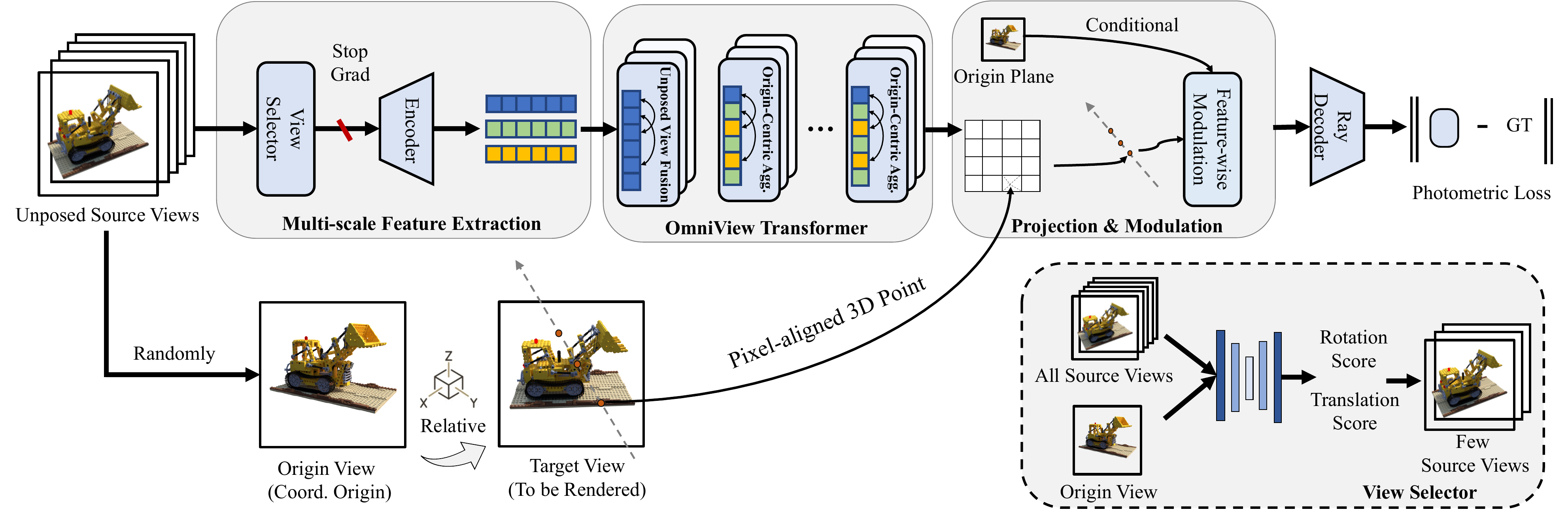}
\vspace{-0.5em}
\caption{\textbf{The overall pipeline of the proposed PF-GRT.} Given unposed source images with a specified \textit{origin view}, PF-GRT selects a limited number of source images closest to the origin view. Multi-scale 2D features are extracted, and the \textit{OmniView Transformer} is used for aggregating features from the unposed sources towards the origin plane. The 3D point feature on the target ray is initialized via projection, and the final pixel intensities are modulated and then decoded using another Transformer.}
\label{fig:arc}
\end{center}
\vspace{-1.5em}
\end{figure*}

\paragraph{Generalizable Neural Scene Representations}
Building generalizable feature volumes dates back to Neural Volumes~\cite{lombardi2019neural}, wherein an encoder-decoder framework is adopted to create a feature volume. Later on, NeRF~\cite{mildenhall2021nerf} and its follow-ups~\cite{barron2021mip,Barron2022MipNeRF3U,yu2021plenoctrees, fridovich2022plenoxels,liu2020neural,hu2022efficientnerf,sun2022direct,muller2022instant,wang2022r2l,garbin2021fastnerf,hedman2021baking,reiser2021kilonerf,attal2021learning, Sitzmann2021LightFN, Feng2021SIGNETEN} have emerged as effective scene representations. However, their costly per-scene fitting nature constitutes a significant limitation. 
Generalizable NeRFs endeavor to circumvent time-consuming optimization by conceptualizing novel view synthesis as a cross-view image-based interpolation problem. 
NeuRay~\cite{liu2022neural}, IBRNet~\cite{wang2021ibrnet}, MVSNeRF~\cite{chen2021mvsnerf}, and PixelNeRF~\cite{yu2021pixelnerf} assemble a generalizable 3D representation using features aggregated from observed views.
GPNR~\cite{suhail2022generalizable} and GNT~\cite{wang2022attention} enhance the novel view renderings with a Transformer-based aggregation process. A view transformer aggregates image features along epipolar lines, while a ray transformer combines coordinate-wise point features along rays through the attention mechanism.
\vspace{-3mm}
\paragraph{Pose-free NeRFs}
Numerous efforts have been exerted to diminish the necessity for calibrated camera poses during NeRF training.
NeRF- -~\cite{wang2021nerf} makes an early endeavor to optimize camera parameters with NeRF training for forward-facing scenes simultaneously.
BARF~\cite{lin2021barf} refines NeRFs from imperfect (or even unknown) camera poses via coarse-to-fine registration, while
GARF~\cite{chng2022garf} incorporates Gaussian activations. NoPe-NeRF~\cite{bian2023nope} employs monocular depth priors to restrict the relative poses between consecutive frames.
Efforts have also been made to expand generalizable NeRF toward unposed images.
PixelNeRF~\cite{yu2021pixelnerf} builds a generalizable feature volume that estimates novel views from single-view observation, which can be unposed.
MonoNeRF~\cite{fu2023mononerf} assumes that the camera motion between adjacent frames is small, and disentangles the predicted depth from the camera pose. It decodes another input frame based on these inputs with a volume rendering representation to formulate the reconstruction loss.
FlowCam~\cite{smith2023flowcam} explicitly estimates the camera pose of the video frame by fitting a 3D scene flow field with the assistance of a pretrained 2D optical flows model.
Generalizable methods, SRT~\cite{sajjadi2022scene} and RUST~\cite{sajjadi2023rust}, infer a set-latent scene representation from a set of unposed images to synthesize novel views. The concurrent work, Leap~\cite{jiang2023leap}, constructs neural volume with feature aggregation, designed for pose-free object-level novel view synthesis.

\vspace{-3mm}
\paragraph{Transformers as Neural Scene Representations}
Transformers are extensively utilized to represent scenes.
IBRNet~\cite{wang2021ibrnet} processes points sampled from rays using MLP and estimates density via a transformer.
NeRFormer~\cite{reizenstein2021common} employs attention modules to craft feature volumes with epipolar geometry constraints.
LFNR~\cite{suhail2022light}, GPNR~\cite{suhail2022generalizable} and GNT~\cite{wang2022attention} introduces a two-stage transformer-based model to accumulate features along epipolar lines and aggregate features along reference views to produce the color of target rays.
SRT~\cite{sajjadi2022scene} and RUST~\cite{sajjadi2023rust} infer a set-latent scene representation via a vision transformer and parameterize light fields by attending into the scene representation for novel view renderings.

\section{Method}

\paragraph{Overview}

Given an unposed image set with known intrinsics $\{ (\Mat{I}_i \in \real^{H \times W \times 3}, \Mat{K}_i \in \real^{3\times3})\}_{i=0}^{N}$ of a scene,
PF-GRT synthesizes new views in a single forward pass.
Specifically, the relative coordinate system is constructed with one unposed view as the origin, and the target view is defined using relative transformation.
The \textit{OmniView Transformer} performs Unposed View Fusion and Origin-Centric Aggregation towards the origin view.
Pixel-aligned 3D point feature on the target ray is initialized by projecting onto the origin plane.
The pixel intensities are subsequently modulated, conditioned on the origin plane, and decoded using another Transformer.
To determine the most suitable source views for aggregation, a view selector is designed to identify the closest $K$ views ($K \leq N$) relative to the origin view, based on global feature distances.
An overview of our pipeline is illustrated in Figure~\ref{fig:arc}.

\subsection{OmniView Transformer}
We propose to use Transformer for learning global feature matching across unposed source views to find the best matching feature for rendering a target ray.

\vspace{-3mm}
\paragraph{Unposed View Fusion}
Having extracted the multi-view feature from a shared 2D encoder, tokens are initialized by subdividing the feature map into $M \times M$ patch grids, which enables generalization to various datasets under different resolutions.
The Unposed View Fusion (UVF) module aggregates the features from each unposed view to capture long-range global context. Specifically, the UVF layer performs self-attention on the token set $\Mat{\mathcal{F}} \in \{\Mat{f}^{1}, ..., \Mat{f}^{ M \times M}\}$ of each unposed source view, which consists of $ M \times M$ discrete tokens:
\begin{align}
\Mat{\mathcal{F}} = \text{FFN}(\text{Attention}(\Mat{\mathcal{F}}, \Mat{\mathcal{F}}))
\end{align}
Here, we utilize attention layers to retrieve relevant information within the given view and update the view-specific feature to be embedded with global context information.
\vspace{-3mm}
\paragraph{Origin-Centric Aggregation}
Next, we propose propagating the multi-view cue from unposed views toward the origin view ($\Mat{I}_0$), which acts as the root of the constructed local coordinate system for rendering. The proposed Origin-Centric Aggregation (OCA) performs the propagation sequentially for each source view via:
\begin{align}
\Mat{\mathcal{F}}_0 = \text{FFN}(\text{Attention}(\Mat{\mathcal{F}}_0, \Mat{\mathcal{F}_i})), \quad \text{i} \in \text{N}
\end{align}
The amalgamation of source features toward the origin view enriches the multi-view information in the updated origin feature plane.
The pseudocode implementation of the OmniView Transformer is provided in the Appendix.

\begin{table*}
\centering
 \resizebox{1.96\columnwidth}{!}{
\begin{tabular}{c|ccc|ccc|ccc|ccc} 
\toprule
\multirow{2}{*}{\textbf{Methods}} & \multicolumn{3}{c|}{Real Forward-facing(LLFF)} & \multicolumn{3}{c|}{RealEstate10K Datasets} & \multicolumn{3}{c|}{Shiny Datasets}
& \multicolumn{3}{c}{NeRF Synthetic Objects}\\ 
\cline{2-13}
                                  & PSNR$\uparrow$ & SSIM$\uparrow$ & LPIPS$\downarrow$ 
                                  & PSNR$\uparrow$ & SSIM$\uparrow$ & LPIPS$\downarrow$ 
                                  & PSNR$\uparrow$ & SSIM$\uparrow$ & LPIPS$\downarrow$
                                   & PSNR$\uparrow$ & SSIM$\uparrow$ & LPIPS$\downarrow$\\ 
\hline
PixelNeRF~\cite{yu2021pixelnerf}   
&8.379 &0.313 &0.643   
& 9.008 &0.407 & 0.503     
&9.025 &0.285 &0.607
& 7.105 & 0.565 & 0.413 
\\
PixelNeRF-ft   
&\cellcolor{top3}10.234 &\cellcolor{top3}0.346 &\cellcolor{top3}0.593
&11.115 &0.462 &0.470
 &\cellcolor{top3}10.130 &\cellcolor{top3}0.304 &\cellcolor{top3}0.555
  &7.561 &0.569 &0.406
          \\
UpSRT~\cite{sajjadi2022scene}    
& \cellcolor{top2}16.669 &\cellcolor{top2}0.541 &\cellcolor{top2}0.418
& \cellcolor{top2}16.833 &\cellcolor{top2}0.593 &\cellcolor{top2}0.314
& \cellcolor{top2}15.117 &\cellcolor{top2}0.471 &\cellcolor{top2}0.428
&  \cellcolor{top3}15.725 &\cellcolor{top3}0.814 &\cellcolor{top3}0.205\\ 
LEAP~\cite{jiang2023leap}   
&  9.220 &0.228 &0.694
& \cellcolor{top3}11.336 &\cellcolor{top3}0.527 &\cellcolor{top3}0.459
& 9.659 &0.257 &0.668
& \cellcolor{top2}18.020 &\cellcolor{top2}0.831 &\cellcolor{top2}0.187
  \\ \hline
Ours          
& \cellcolor{top1}{22.728} & \cellcolor{top1}{0.778} & \cellcolor{top1}{0.180}

& \cellcolor{top1}{24.208} & \cellcolor{top1}{0.789} & \cellcolor{top1}{0.190}

& \cellcolor{top1}19.192 &\cellcolor{top1}0.604 &\cellcolor{top1}0.338 
& \cellcolor{top1}{22.832} & \cellcolor{top1}{0.835} &\cellcolor{top1}{0.134} 
         \\
         \hline \bottomrule
\end{tabular}}
\caption{\textbf{Quantitative Comparison in a Generalizable Pose-Free Setting.}
PF-GRT outperforms previous pose-free methods that utilize both single-view feature volume (PixelNeRF)\cite{yu2021pixelnerf} and multi-view ``set of latents''(UpSRT)\cite{sajjadi2022scene}, in addition to aggregation to neural volume (LEAP)~\cite{jiang2023leap}. Owing to the effective \textit{OmniView Transformer} and the \textit{IBR} formulation, our method can generate novel views with the highest quality.
We color each row as \textbf{\colorbox[RGB]{255,179,179}{best}}, \textbf{\colorbox[RGB]{255,217,179}{second best}},
and \textbf{\colorbox[RGB]{255,255,179}{third best}}. }
\label{tab:main1}
\end{table*}

\begin{figure*}[ht!]
\vspace{-1em}
\vskip 0.2in
\begin{center}
    \centering
    \includegraphics[width=0.79\linewidth]{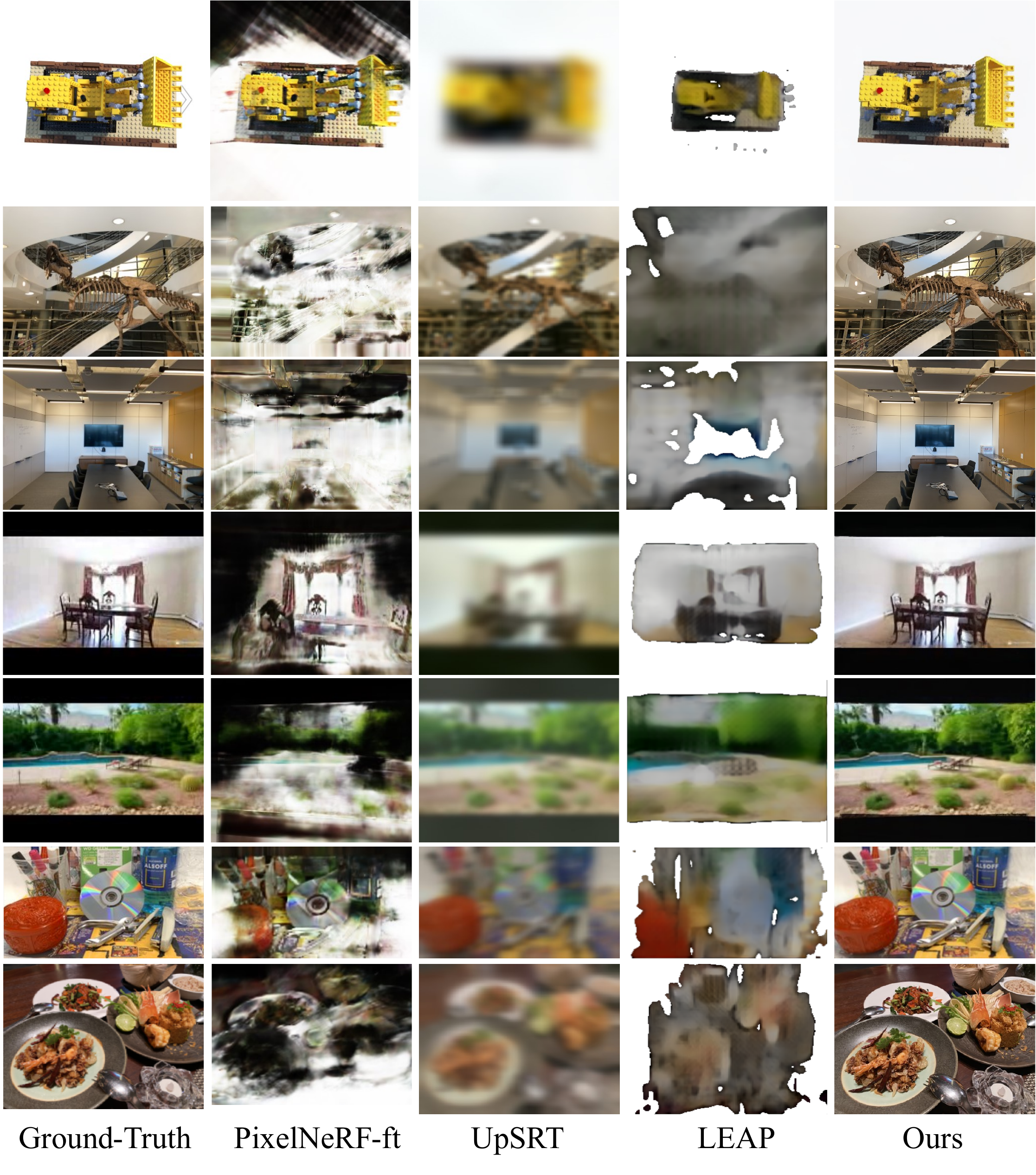} 
\vspace{-4mm}
\caption{\textbf{Qualitative Comparison Under a Generalizable Pose-Free Setting.}
Single-view PixelNeRF~\cite{yu2021pixelnerf} introduces artifacts within the constructed feature volume and shows limited capacity in handling complex scenes.
Multi-view SRT~\cite{sajjadi2022scene} fails to render sharp details in scenes with rich textures.
LEAP~\cite{jiang2023leap} can generalize to object-level datasets, but it fails to scale up to scene-level synthesis.
In contrast, our method more accurately recovers fine details through global correspondence matching and image-based rendering.}
\label{fig:comp_main}
\end{center}
\vspace{-1.3em}
\end{figure*}

\subsection{Target Ray Initialization and Decoding}
To render a pixel $\Mat{c} = (\Mat{o}, \Mat{d})$ in target view, the relative transformation from origin plane is specified. Pixel-aligned 3D points $\Mat{p} \in \{p_1, p_2, ..., p_M\}$ are uniformly sampled between the near and far plane. Point feature are tokenized, modulated and decoded into pixel intensities.

\vspace{-3mm}
\paragraph{Tokenizing 3D Points on Target Ray}
The uniformly sampled points, $\Mat{p}$ are projected onto the feature maps of the origin view, culminating in the initialized point features:
\begin{align}
\hat{\Mat{p}} &= \Mat{K}_0(\Mat{R}_0(\Mat{p}) + t_0) \\
\Mat{F}(\Mat{\mathcal{F}_0}, \hat{\Mat{p}}) &= \operatorname{Interpolation}(\Pi(\hat{\Mat{p}}, \Mat{\mathcal{F}_0}))
\end{align}
Here, $\Mat{R}_0$ and $t_0$ symbolize the relative rotation and translation between the target view to the origin view, $\Pi$ represents the projection function, and $\Mat{\mathcal{F}}_0$ is the extracted feature maps on origin view, formulating the pixel-aligned 3D point-wise feature $\hat{\Mat{p}} \in \{\hat{p_1}, \hat{p_2}, ..., \hat{p_M\}}$. 
\vspace{-3mm}
\paragraph{Origin-Conditioned Modulation Layer}
When projecting points into regions of the scene that are obscured from the camera's viewpoint, or where erroneous feature matchings from unposed source views occur, the projected points often become occluded or missed, degrading the rendered quality. To counteract this issue, an origin view-conditioned layer is introduced to modulate the point features through an affine transformation, utilizing the extracted global coherence derived from the origin view.
Formally, a Feature-wise Linear Modulation (FiLM) layer~\cite{perez2018film} is utilized to scale and shift the point feature $\hat{\Mat{p}}$ following:
\begin{align}
\gamma^\tau,  \beta^\tau &= \operatorname{MLP}_{\gamma}(\operatorname{GAP}(\Mat{\mathcal{F}}_0)),   \operatorname{MLP}_{\beta}(\operatorname{GAP}(\Mat{\mathcal{F}}_0)) \\
\hat{\Mat{p}} &:= \gamma^\tau \hat{\Mat{p}} + \beta^\tau
\end{align}
This modulation is formulated by two groups of parameters, $\gamma^\tau$ and $\beta^\tau$, resulting in the modulated point-wise feature $\hat{\Mat{p}}$. $\operatorname{GAP}$ denotes Global Average Pooling.

\vspace{-3mm}
\paragraph{Ray-based Rendering with Transformers}
Recent research advocates the utilization of the Transformer architecture to adaptively learn the blending weights along the ray for each point, augmenting both expressiveness~\cite{wang2022attention} and generalization~\cite{suhail2022generalizable}. Instead of using the simplified version of volume rendering~\cite{max1995optical} from NeRF~\cite{mildenhall2021nerf}, we employ such an ``attention'' mechanism to determine the aggregation weights for each sampled point feature in a data-driven way, to decode the final pixel intensity:
\begin{align}
   \Mat{c}(\Mat{o}, \Mat{d}) = \text{MLP} \circ \text{Mean} \circ \text{Attention}(\hat{\Mat{p}}, \hat{\Mat{p}})),
\end{align}

\subsection{View Selection via Feature Similarity}
View selection aims to select efficiently a few source images, which is the nearest to the origin view, to reduce the computational redundancy when performing OmniView attention.
Specifically, a network is designed to extract multi-scale features~\cite{he2016deep} from all source images, and multiple decoding heads are devised for regressing the relative rotation and translation scores between $\Mat{I}_0$ and each source image $\{\Mat{I}_i, i \neq 0\}$.
In particular, four decoding heads are utilized for estimating the three normalized relative angles and the distance value between the two images. Top $K$ images are selected out of the $N$ ($K \leq N$).

\subsection{Training and Inference}
During the training phase, the view selector identifies the nearest $K$ source images from the $N$ unposed source images. This selection procedure is guided by a specified loss function that operates based on the online-computed relative angle and distance values of each image pair. 
\vspace{-1mm}
\begin{align} \label{eqn:selector_loss}
\Theta_s^* = \arg\min_{\Mat{\Theta}} (\left\lVert \angle(\Mat{I}_0, \Mat{I}_i) - \angle_{gt} \right\rVert_2^2 + \left\lVert d(\Mat{I}_0, \Mat{I}_i) - d_{gt} \right\rVert_2^2 ) .
\end{align}
The rest part of the model is optimized utilizing the $\mathcal{L}_2$ distance between the rendered target pixels and the corresponding ground-truth pixels, as exemplified by:
\vspace{-1mm}
\begin{align} \label{eqn:photo_loss}
\Theta_t^* = \arg\min_{\Mat{\Theta}} \left\lVert \Mat{C}(\Mat{r}_{i} \vert \Mat{\Theta}, \Mat{\theta}, \Mat{x}) - \Mat{C}_{gt} \right\rVert_2^2.
\end{align}
Note that the training requires \underline{ground-truth camera poses} for calculating the score for the view selector, and the absolute poses are converted into relative poses for supervising the model training.
For inference on unseen scenes, there is \underline{no reliance on any pose estimator}. Instead, target views are specified based on a relative transformation from the origin view, which can be selected arbitrarily. Other source views are determined by the view selector.

\section{Experiments}

\subsection{Implementation Details}
\paragraph{Datasets.}
We train PF-GRT on a large-scale dataset, spanning from $\sim$ 1000 Google Scanned Objects~\cite{downs2022google} to real-world datasets: sparsely-sampled RealEstate10K~\cite{zhou2018stereo}, 100 scenes from the Spaces datasets~\cite{flynn2019deepview}, and 102 real-world scenes captured using handheld phones~\cite{mildenhall2019local,wang2021ibrnet}.
We evaluate the zero-shot accuracy on diverse datasets without finetuning:
\begin{itemize}
\item \textbf{LLFF Datasets}\cite{mildenhall2019local} contain eight real-world forward-facing scenes, with a resolution of 504 $\times$ 378.
\item \textbf{RealEstate10K}\cite{zhou2018stereo}. The test set consists of four real-world scenes at a resolution of 176$\times$144.
\item \textbf{Shiny Datasets}\cite{wizadwongsa2021nex} contain eight scenes with challenging lighting conditions, with evaluations conducted on 8$\times$ downsampled images.
\item \textbf{Blender Datasets}, the widely-adopted synthetic data created by NeRF\cite{mildenhall2021nerf}, encompasses eight objects with a tested resolution of 400$\times$400.
\end{itemize}

\vspace{-3mm}
\paragraph{Training and Inference Details.}
PF-GRT is trained end-to-end, with the gradient stopping operation after the view selector. The Adam optimizer is employed to minimize the training loss for the model. The learning rate decreases exponentially over training steps with a base of 10$^{-3}$. The comprehensive training encompasses 250,000 steps, with 4,096 randomly sampled rays during each iteration. In both training and inference phases, 128 points are uniformly sampled along each target ray. Grid number $M$ is set as 7 to balance efficiency and accuracy.

\begin{table*}
\centering
\vspace{-3mm}
 \resizebox{1.49\columnwidth}{!}{
\begin{tabular}{c|ccc|ccc} 
\toprule
\multirow{2}{*}{\textbf{Methods}} & \multicolumn{3}{c|}{Real Forward-facing(LLFF)} & \multicolumn{3}{c}{NeRF Synthetic Objects} \\ 
\cline{2-7}
                                  & PSNR $\uparrow$ & SSIM$\uparrow$ & LPIPS$\downarrow$   & PSNR$\uparrow$ & SSIM$\uparrow$ & LPIPS$\uparrow$         \\ 
\hline
IBRNet~\cite{wang2021ibrnet}            & \cellcolor{top3}21.395 & \cellcolor{top2}0.686 &\cellcolor{top2}0.303  &20.027 & 0.813 & 0.145        \\
NeuRay~\cite{liu2022neural}            & \cellcolor{top2}21.520 & 0.681 & \cellcolor{top2}0.303  & \cellcolor{top2}21.424 & \cellcolor{top2}0.832 & \cellcolor{top2}0.135     \\
GNT~\cite{wang2022attention}               & 21.341 & \cellcolor{top3}0.682 &\cellcolor{top3} 0.307   & \cellcolor{top3}20.554 & \cellcolor{top3}0.830 & \cellcolor{top3}0.139
  \\ \hline
Ours              & \cellcolor{top1}{22.728} & \cellcolor{top1}{0.778} & \cellcolor{top1}{0.180} & \cellcolor{top1}{22.832} & \cellcolor{top1}{0.835} & \cellcolor{top1}{0.134}        \\ \hline
\bottomrule
\end{tabular}}
\caption{\textbf{Quantitative Comparison of Robustness to Noisy Poses in source views.} The table presents a performance comparison between PF-GRT and various generalizable NeRFs using the NeRF-Synthetic~\cite{mildenhall2021nerf} and LLFF datasets~\cite{mildenhall2019local}, where both rotation and translation matrices are perturbed with $\sigma$ = 0.003. PF-GRT showcases its robustness in handling pose perturbations in rendered views.
We color each row as \textbf{\colorbox[RGB]{255,179,179}{best}}, \textbf{\colorbox[RGB]{255,217,179}{second best}},
and \textbf{\colorbox[RGB]{255,255,179}{third best}}.}
\label{tab:main2}
\vspace{-3mm}
\end{table*}

\begin{figure*}[t]
\vspace{-1em}
\vskip 0.2in
\begin{center}
    \centering
    \includegraphics[width=0.99\linewidth]{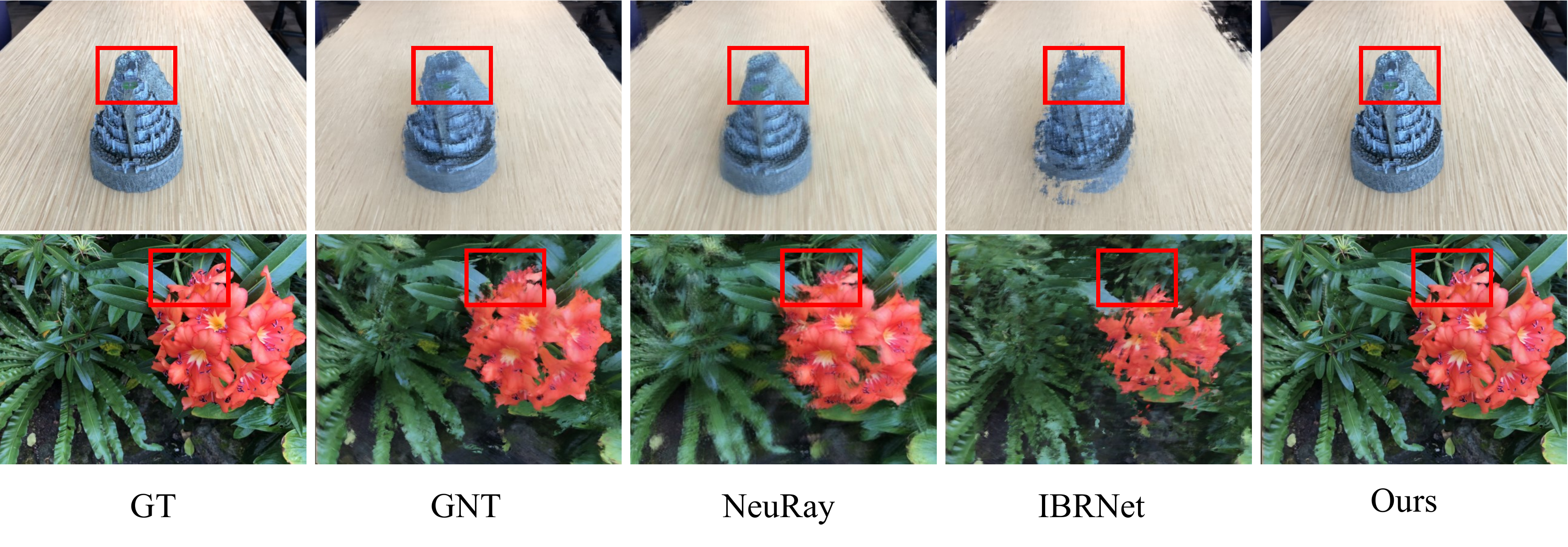}
\vspace{-3mm}
\caption{\textbf{Visualizations of Different Methods Against Noisy Poses on Source Images When Rendering}. All adopted generalizable NeRFs suffer from noisy camera poses in source views at evaluation, even with very mild perturbation (e.g., $\sigma$=0.003).}
\label{fig1:robustness}
\end{center}
\vspace{-1.3em}
\end{figure*}

\begin{figure}[t]
\vspace{-1.0em}
\vskip 0.2in
\begin{center}
    \centering
    \includegraphics[width=1 \linewidth]{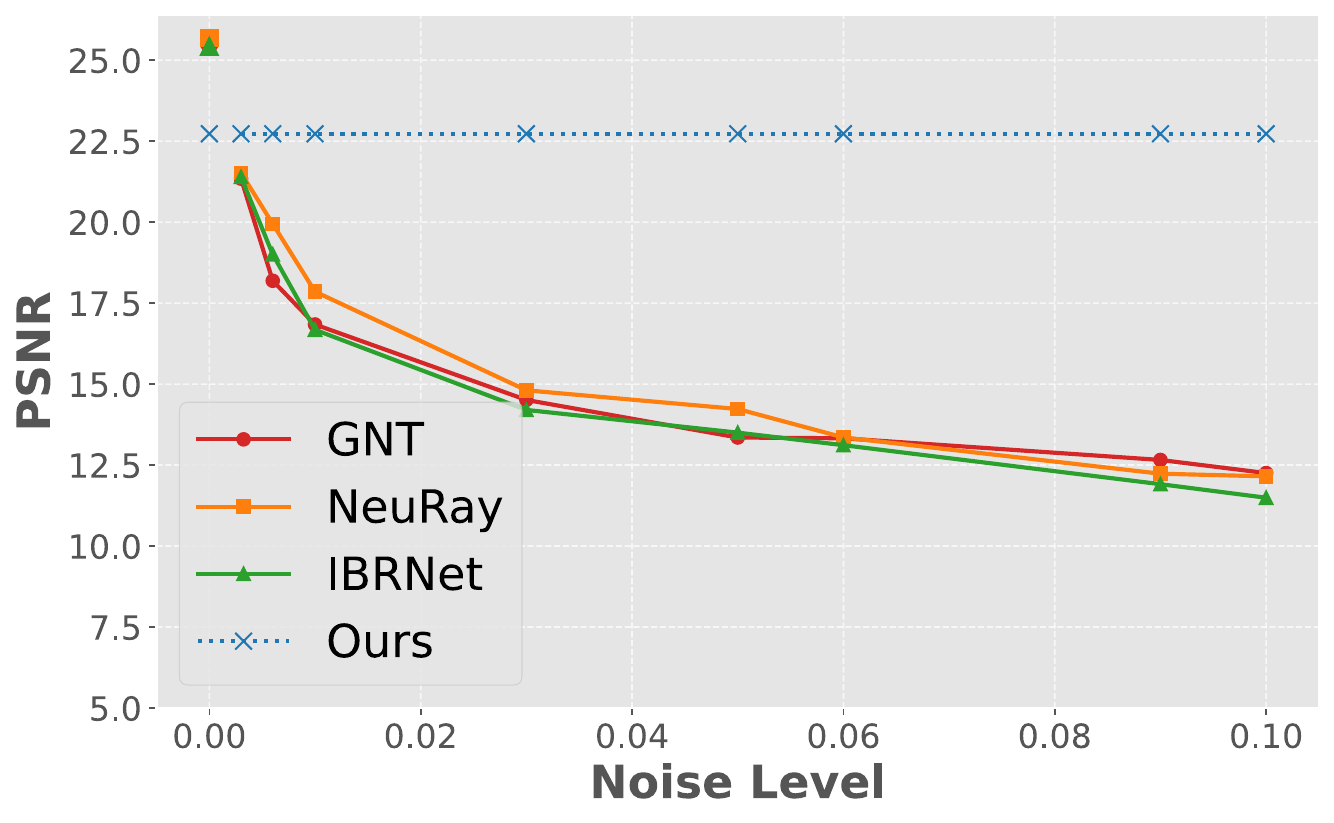}
\vspace{-3mm}
\caption{\textbf{Visualizations of generalizable NeRFs with different noise level}. We increases the noise from ($\sigma$=0.003) to ($\sigma$=0.1), methods that rely on camera poses for cross-view aggregation are decreasing in rendering quality. Our method demonstrates the robustness against test pose noises, as we perform global feature matching toward the origin view, instead of using epipolar constraints.}
\label{fig1:different_noise}
\end{center}
\vspace{-1.3em}
\end{figure}

\vspace{-3mm}
\paragraph{Baselines.}
We evaluate pose-free novel view synthesis with UpSRT~\cite{sajjadi2022scene}, single-view based PixelNeRF~\cite{yu2021pixelnerf}, and the generalizable LEAP~\cite{jiang2023leap}. We re-trained UpSRT~\cite{sajjadi2022scene}\footnote{\url{https://github.com/stelzner/srt}} on the same datasets as ours for 400K iterations until fully converged. We report both the vanilla PixelNeRF from their provided checkpoint, and the finetuned version on the same dataset as ours. We use the checkpoint from LEAP\cite{jiang2023leap} trained on OmniObject datasets~\cite{wu2023omniobject3d}.
We also test robustness against camera pose noises in source views with other generalizable NeRFs~\cite{wang2021ibrnet, wang2022attention,liu2022neural}: we synthetically perturb the camera poses with additive Gaussian noise on both rotation and translation vectors, akin to BARF~\cite{lin2021barf}.

\vspace{-3mm}
\paragraph{Metrics}
The rendering quality is reported using metrics such as peak signal-to-noise ratio (PSNR), structural similarity (SSIM), and perceptual similarity via LPIPS~\cite{zhang2018unreasonable}.

\subsection{Generalizable Pose-free Novel View Synthesis.}
Table~\ref{tab:main1} showcases that PF-GRT surpasses the best-performing multi-view pose-free SRT~\cite{sajjadi2022scene}, LEAP~\cite{jiang2023leap}, and single-view PixelNeRF on all unseen test scenes, encompassing synthetic datasets, forward-facing datasets, RealEstate10K datasets and Shiny datasets with complex lighting conditions. 
This validates the efficacy of the designed OmniView Transformer for effective view aggregation under pose-free setting, and the image-based rendering to generate detailed textures. 
We present qualitative results in Figure~\ref{fig:comp_main}, wherein it can be observed that the ``latent'' representation in SRT overlooks the image details, and PixelNeRF struggles under complex scenes using single-view feature volume-based neural rendering. 
LEAP has difficulties to generalize to scene-level test cases.
See the video in supplementary materials for detailed comparisons.

\subsection{Robustness against Noisy Poses.}
Multi-view images captured in the real world typically require a pre-processing step (e.g., COLMAP~\cite{schoenberger2016sfm}) to compute the poses. However, this computation is slow, especially when the number of source images is large, and often contains errors~\cite{lin2021barf}. We examine the current best-performing generalizable NeRFs against noisy camera poses in the tested source views, a practical concern. Following~\cite{lin2021barf}, who apply additive Gaussian perturbation at different levels to the camera poses, we directly test the trained generalizable model with the provided checkpoint to assess robustness. It is evident from Figure~\ref{fig1:robustness} that all generalizable methods suffer from noisy camera poses, with significant degradation in performance even under a small amount of noisy calibration ($\sigma$=0.003).
On the other hand, our framework PF-GRT, which generates new views in a feed-forward pass without estimating camera poses, demonstrates stability in rendering quality. This stability is attributed to our approach of not relying on absolute camera poses for cross-view feature aggregation but instead learning the cross-view feature mapping from large-scale training using the OmniView Transformer. Consequently, the source view pose noises do not affect the rendering quality.

Figure~\ref{fig1:different_noise} visualizes the effect of different noise levels on source views in the evaluation, illustrating that subtle noise significantly decreases the rendering quality.
Quantitative results, with a noise level of 0.003 on both real forward-facing and synthetic objects datasets, are presented in Table~\ref{tab:main2}.

 
\subsection{Ablation Study}
\begin{figure*}[t]
\vspace{-1em}
\vskip 0.2in
\begin{center}
    \centering
    \includegraphics[width=0.9\linewidth]{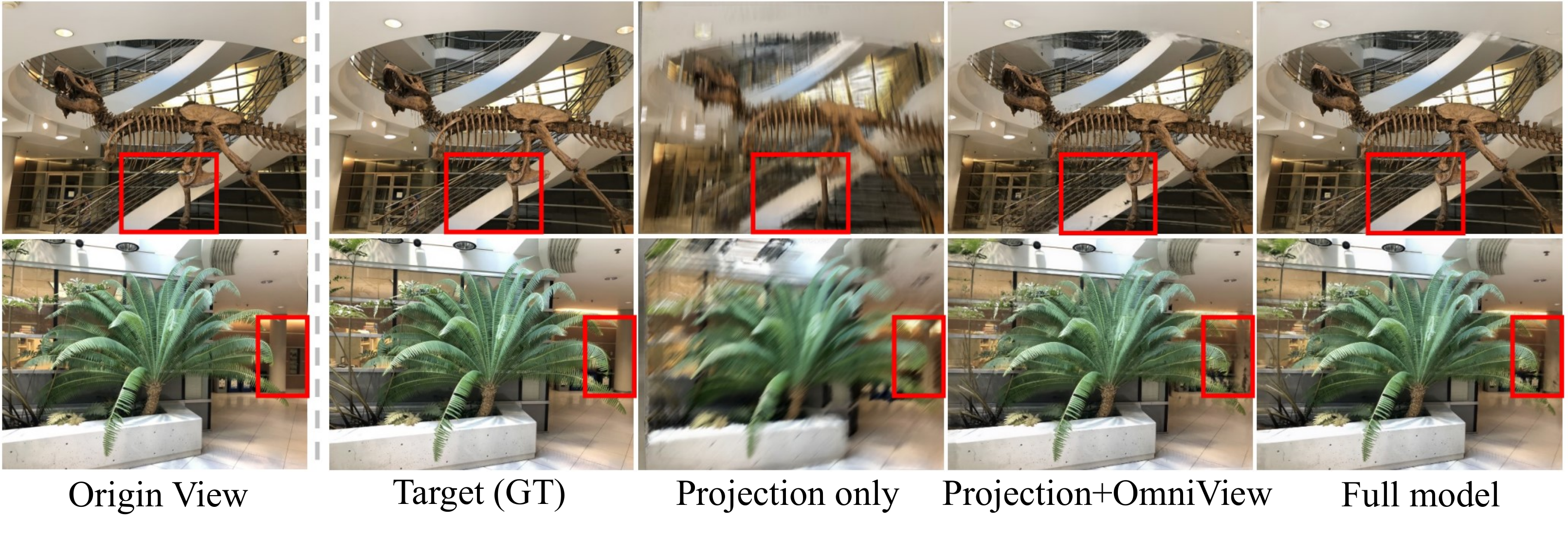} 
\vspace{-3mm}
\caption{\textbf{Visualization for Ablation Study.} We visualize the ``origin view'' (1st column), ``ground-truth target view'' (2nd column), rendered images using projection and ray attention (3rd column), the incorporation of the \textit{OmniView Transformer} (4th column), and our full model (last column).}
\label{fig:ablation}
\end{center}
\vspace{-1.3em}
\end{figure*}

We now execute a series of ablation studies regarding our module choice on the LLFF datasets~\cite{mildenhall2019local} and average the metrics across all scenes. The evaluation begins with the use of the ``origin view" directly for evaluation, and subsequently, we incrementally integrate the proposed techniques (\textit{Pixel-aligned Feature}, \textit{OmniView Transformer}, \textit{Conditional Modulation Layer}) in this study.
\vspace{-3mm}
\paragraph{Pixel-aligned Feature?}
We study utilizing the relative coordinate system between the origin view and the target views, and obtain the 3D point feature along the target ray by directly projecting onto origin feature plane. 
The final pixel intensities of target ray are regressed using ray attention.
As we can see in the second row of Table~\ref{table:ablation_main} and the third column of Figure~\ref{fig:ablation}, the missing of multi-view cues results in an inferior rendering quality.

\begin{table}[t!]
\small
\centering
\vspace{-3mm}
\resizebox{0.99\columnwidth}{!}{
\begin{tabular}{@{}cccccc@{}}
\toprule
  Pixel-aligned   & OmniView   &  Modulation
 & \multirow{2}{*}{PSNR$\uparrow$}  & \multirow{2}{*}{SSIM$\uparrow$}  & \multirow{2}{*}{LPIPS$\downarrow$}  \\
  Feature  & Transformer & FiLM-Layer &   &     \\  
  \midrule
 \xmark & \xmark & \xmark  &  14.198     &0.355    &0.407  \\
 \cmark & \xmark & \xmark &  16.519     &0.412    &0.400  \\
 \cmark & \cmark & \xmark & 22.287 & 0.740 &  0.197 
 \\
\cmark & \cmark & \cmark & \bf{22.728} &\bf{0.778} &\bf{0.180} \\
  \hline \bottomrule
\end{tabular}}
\caption{{\bf Ablation study of the proposed components in PF-GRT.}
We start by constructing the relative coordinate system between origin and target view, utilizing pixel-aligned feature (2nd row) shows a better metrics than the baseline that directly evaluating the origin view (1st row).
The introduction of OmniView Transformer (3rd row) significantly improve the PSNR from 16.519dB to 22.287dB.
Additionally, the feature-wise modulation layer further improve the PSNR to 22.728dB (last row).}\label{table:ablation_main}
\end{table}

\begin{table}[h]
\small
\centering
\vspace{-3mm}
\resizebox{0.99\columnwidth}{!}{
\begin{tabular}{@{}cccccc@{}}
\toprule\toprule
 Random?/GT?/ & Disentangled  
 & \multirow{2}{*}{PSNR$\uparrow$}  & \multirow{2}{*}{SSIM$\uparrow$}  & \multirow{2}{*}{LPIPS$\downarrow$}  & \multirow{2}{*}{Select Acc.$\uparrow$}  \\
 View Selector? & Decoder Heads   &   &   &  &  \\  \midrule
Random & N/A  &  17.841 & 0.505  &  0.390   &  0.017      \\
View Selector & \xmark &   22.243 &0.756 &0.198  &  0.688   \\
View Selector & \cmark & 22.728 &0.778 &0.180  &  0.731
    \\
GT & N/A & 24.275 & 0.822 & 0.135   &  1.000      \\
  \bottomrule\bottomrule
\end{tabular}}
\caption{\textbf{Ablation on the View Selector.} The incorporation of the view selector finds the best matched source images, enabling PF-GRT to effectively aggregate cross-view features. The employment of disentangled decoder heads for each axis of rotation and translation vector further improves the view selection accuracy and, thus, the synthesis quality. PF-GRT, which utilizes \underline{ground-truth poses} for view selection, is listed in the last row.}\label{table:ablation_view_selector}
\end{table}

\vspace{-4mm}
\paragraph{OmniView Transformer?}
We further employ the \textit{OmniView Transformer} for Unposed-View Fusion and Origin-Centric Aggregation, using a data-driven attention mechanism focused on the origin view.
This design significantly improves quantitative metrics, with an increase in PSNR from 16.519 to 22.287, as shown in the third row of Table~\ref{table:ablation_main}.
Additionally, Figure~\ref{fig:ablation} demonstrates that the visual quality is also improved by integrating missing information from other source views.
\vspace{-4mm}
\paragraph{Conditional Modulation Layer?}
Conditioned on the statistics of the origin view, the 3D point-wise features on the target ray are affine-transformed using learnable parameters. This adjustment aids in filling in missing regions in occluded areas (see the second-last column of Figure~\ref{fig:ablation}) especially when cross-view aggregation is imprecise and the modulation layer can guide the projection toward a plausible solution.
The last row of Table~\ref{table:ablation_main} shows that the modulation layer improves SSIM from 0.74 to 0.778.
\vspace{-3mm}

\paragraph{Analysis on the Viewpoint Selector.}
Initially, we assess the use of random source view selection in the evaluation, where the selected source views may not be ideal for feature aggregation, resulting in a 0.505 SSIM metric (1st row of Table~\ref{table:ablation_view_selector}). Utilizing the selector to regress the averaged relative viewing direction and distance directly elevates the SSIM to 0.756, a significant improvement. Employing separate decoder heads to disentangle the similarity score for the three axes of relative rotation and translation distance further enhances the quantitative metrics to 0.778 ($\uparrow$ 0.02 in SSIM). Additionally, we illustrate the use of ground-truth poses to identify the nearest source views with viewing directions most analogous to the ``origin view,'' serving as the upper bound of the view selector.

\clearpage
\setcounter{page}{1}
\maketitlesupplementary

\renewcommand{\thepage}{A\arabic{page}}  
\renewcommand{\thesection}{A\arabic{section}}   
\renewcommand{\thetable}{A\arabic{table}}   
\renewcommand{\thefigure}{A\arabic{figure}}

\section{Motivation of OmniView Transformer}
\begin{figure*}[ht!]
\vspace{-1em}
\vskip 0.2in
\begin{center}
    \centering
    \includegraphics[width=0.99\linewidth]{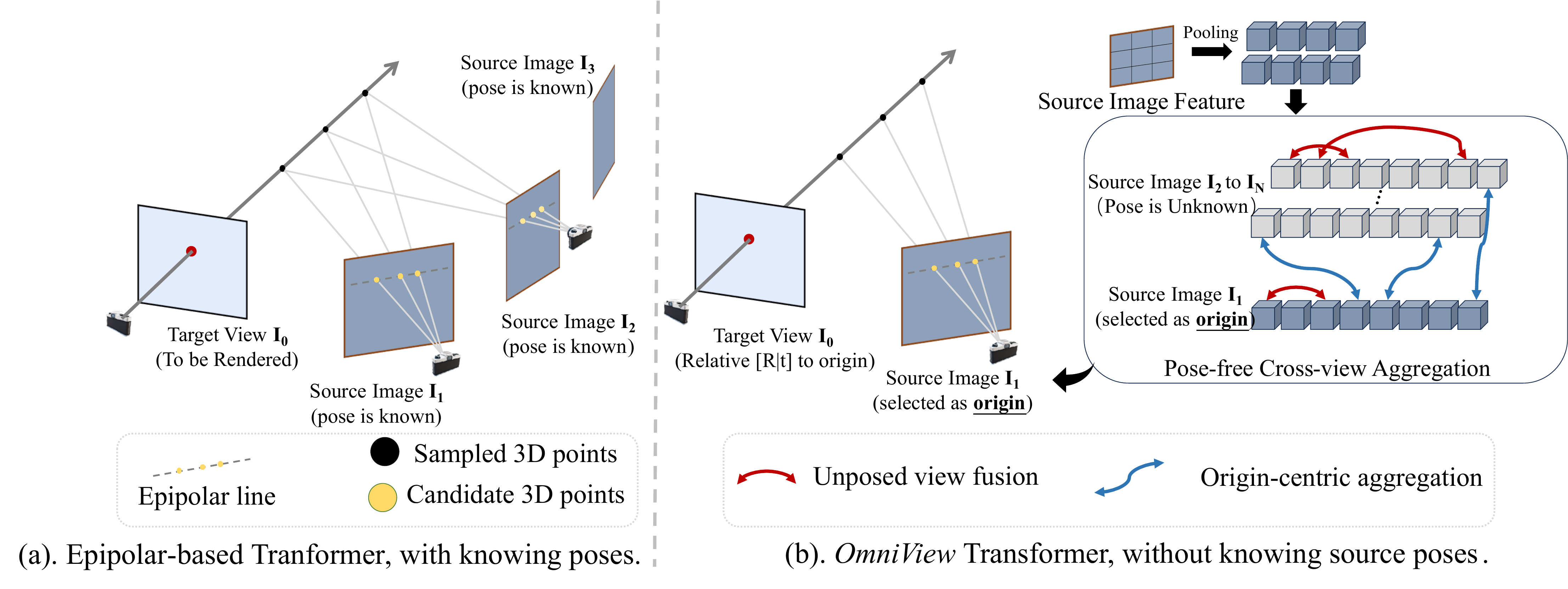} 
\caption{Illustration of \textit{Epipolar Attention} and \textit{OmniView Attention}. The figure is the same with Figure 2 in the main draft.}
\label{fig1:supp-ill}
\end{center}
\vspace{-1.3em}
\end{figure*}

As previously mentioned in the main paper, knowing the multi-view camera poses enables the framework design to search for correspondences along the epipolar line. As illustrated in Figure~\ref{fig1:supp-ill}(a), we consider a simple case with several source views. For the pixel to be rendered in the target view, epipolar attention used in~\citep{wang2022attention,suhail2022generalizable} builds correspondences among the target ray and epipolar line of neighboring source images.
However, without knowing the poses, we are unable to build such a search pattern, and thereby, we resort to utilizing an attention mechanism to search over all source pixels toward the origin view (the relative coordinate system is constructed), which is the origin of the relative coordinate system. We propose the use of a CNN network to extract multi-scale feature maps.
Subsequent to the CNN encoder, these extracted feature maps from source views ${ (\Mat{I}i \in \real^{H \times W \times 3})}{i=0}^{K}$ are subdivided into $M \times M$ grids within each view, facilitating the model to be agnostic to diverse image resolutions (as shown in Figure~\ref{fig1:supp-ill} (b)).
The \textit{Unposed View Fusion}, which performs intra-image long-range global context aggregation, is designed to retrieve relevant information within the source views, while the \textit{Origin-Centric Aggregation} focuses on capturing cross-relationships across these two views. This way, inter-image feature interaction between images is facilitated.

\section{Implementation Details}
\paragraph{Memory-Efficient OmniView Transformer}
The most straightforward method for aggregating the initially projected 3D point feature involves building cross-attention between the target 3D point feature and all source pixels. However, this approach is intractable as it cannot scale to high-resolution input images and a large number of source views. Therefore, we propose to leverage the 8$\times$ downsampled CNN features and pool them into a fixed number of 2D grids (here, we use a 7$\times$7 grid) for each view. Consequently, our design is agnostic to input resolution, allowing attention to be performed in a patch-wise manner. Nevertheless, during training, the sampled ray is typically large (e.g., 4096 in PF-GRT), incurring 4096 $\times$ 128 sampled points in each iteration. The cross-attention among sampled points and tokenized patches in the source views remains intractable. Therefore, akin to the [CLS] token in Vision Transformer~\citep{dosovitskiy2020image}, we employ the cross-attention mechanism to propagate multi-view information in source views toward the origin view. We then project the sampled 3D points onto the starting view, ensuring an efficient implementation regardless of the number of source views used. Please refer to the PyTorch-like pseudo-code Algorithm~\ref{alg:omniview} for a detailed explanation.
%

\begin{algorithm}
\caption{OmniView Transformer: PyTorch-like Pseudocode}\label{alg:omniview}
\begin{algorithmic}
\State $\Mat{p}_{t} \rightarrow \text{points coordinate in target view} (N_{\text{rays}}, N_{\text{pts}}, 3)$
\State $\Mat{X}_{0} \rightarrow \text{flattened tokens in origin} (1, N_{\text{patch}}, C)$
\State $\{\Mat{X}_{i}\}_{i=1}^{K} \rightarrow \text{flattened tokens in source views} (K, N_{\text{patch}}, C)$
\State $\Mat{f}_{t} \rightarrow \text{projected points feature} (N_{\text{rays}}, N_{\text{pts}}, D)$
\State $f_{Q}, f_{K}, f_{V}, f_{\text{rgb}} \rightarrow \text{functions that parameterize MLP layers}$
\\
\For {$0 \leq i \leq K$}  \Comment{Unposed View Fusion}
\State $\Mat{Q} = f_{Q}(\Mat{X_{i}})$
\State $\Mat{K} = f_{K}(\Mat{X_{i}})$
\State $\Mat{V} = f_{V}(\Mat{X_{i}})$
\State $\Mat{A} = \operatorname{matmul}(\Mat{Q}, \Mat{K}^{T}) / \sqrt{D}$
\State $\Mat{A} = \operatorname{softmax}(\Mat{A}, \operatorname{dim}=-1)$
\State $\Mat{X_{i}} = \operatorname{matmul}(\Mat{A}, \Mat{V})$
\EndFor

\\
\For {$1 \leq i \leq K$}  \Comment{Origin-Centric Aggregation}
\State $\Mat{Q} = f_{Q}(\Mat{X_{i}})$
\State $\Mat{K} = f_{K}(\Mat{X_{0}})$
\State $\Mat{V} = f_{V}(\Mat{X_{0}})$
\State $\Mat{A} = \operatorname{matmul}(\Mat{Q}, \Mat{K}^{T}) / \sqrt{D}$
\State $\Mat{A} = \operatorname{softmax}(\Mat{A}, \operatorname{dim}=-1)$
\State $\Mat{X_{0}} = \operatorname{matmul}(\Mat{A}, \Mat{V})$
\EndFor

\\
\For {$0 \leq i \leq (N_{\text{rays}} \times N_{\text{pts}})$} \Comment{Point-wise projection}
\State $\Mat{f}_t^i = \operatorname{interp.}(\operatorname{proj.}(\operatorname{modulation}(\Mat{p}_t^i), \Mat{X}_0))$
\EndFor

\\
\For {$0 \leq i \leq N_{\text{rays}}$}  \Comment{Ray attention}
\State $\Mat{Q} = f_{Q}(\Mat{f}_t^i)$
\State $\Mat{K} = f_{K}(\Mat{f}_t^i)$
\State $\Mat{V} = f_{V}(\Mat{f}_t^i)$
\State $\Mat{A} = \operatorname{matmul}(\Mat{Q}, \Mat{K}^{T}) / \sqrt{D}$
\State $\Mat{A} = \operatorname{softmax}(\Mat{A}, \operatorname{dim}=-1)$
\State $\Mat{f}_t^i = \operatorname{matmul}(\Mat{A}, \Mat{V})$
\EndFor

\\
\State $\text{RGB} = f_{\text{rgb}}(\operatorname{mean}_{i=1}^{N_{\text{pts}}}(\Mat{f}_t^i))$

\end{algorithmic}
\end{algorithm}

\section{Additional Experiments}
\paragraph{Scene-wise Quantitative Metrics}
Table~\ref{tab:llff_breakdown}, Table~\ref{tab:shiny_breakdown} Table~\ref{tab:real-estate_breakdown}, and able~\ref{tab:blender_breakdown} include a scene-wise quantitative results presented in the main paper. Our method quantitatively surpasses both the generalizable single-view based method PixelNeRF~\citep{yu2021pixelnerf} (and PixelNeRF with finetuning) and multi-view based method UpSRT~\citep{sajjadi2022scene} trained with 4000k iterations and LEAP~\cite{jiang2023leap} from the pre-trained model weights. We also include videos to demonstrate our results in the attached video.

\begin{table*}
  \caption{Comparison of PF-GRT with other pose-free generalizable novel view-synthesis methods on the forward-facing LLFF datasets (scene-wise).}
  \label{tab:llff_breakdown}
  \begin{subtable}[t]{\textwidth}
  \centering
  \resizebox{0.9\columnwidth}{!}{
  \begin{tabular}{lcccccccc}
    \toprule
    Models & Trex  & Fern  & Flower  & Leaves  & Room  & Fortress  & Horns  & Orchids   \\
    \midrule
 pixelNeRF & 8.266 & 8.655 & 8.234 & 7.026 & 8.872 & 10.55 & 7.743 & 7.177 \\
pixelNeRF-ft & 9.6914 & 9.541 & 11.9751 & 7.8957 & 11.7662 & 12.729 & 9.7231 & 9.2533 \\
 SRT & 16.383 & 16.7918 & 17.0056 & 14.0108 & 18.3335 & 19.7896 & 16.6579 & 14.3807 \\
 LEAP & 8.6914 & 8.541 & 10.9751 & 6.8957 & 10.7662 & 11.729 & 8.7231 & 8.2533 \\
   \midrule
PF-GRT & 21.489 & 21.847 & 22.786 & 17.725 & 26.836 & 27.261 & 23.866 & 16.139 \\
    \bottomrule
  \end{tabular}}
  \caption{PSNR$\uparrow$}
  \end{subtable}
  \begin{subtable}[t]{\textwidth}
  \centering
  \resizebox{0.9\columnwidth}{!}{
  \begin{tabular}{lcccccccc}
    \toprule
    Models & Trex  & Fern  & Flower  & Leaves  & Room  & Fortress  & Horns  & Orchids   \\
    \midrule
pixelNeRF & 0.351 & 0.326 & 0.24 & 0.127 & 0.492 & 0.418 & 0.275 & 0.161 \\
pixelNeRF-ft & 0.3762 & 0.3639 & 0.3551 & 0.1757 & 0.4983 & 0.5289 & 0.3719 & 0.222 \\
SRT & 0.8166 & 0.7976 & 0.8637 & 0.801 & 0.7821 & 0.8849 & 0.6413 & 0.7349 \\
LEAP & 0.2596 & 0.3175 & 0.3435 & 0.4334 & 0.1388 & 0.2476 & 0.352 & 0.4128 \\
\midrule
PF-GTR & 0.798 & 0.737 & 0.773 & 0.674 & 0.848 & 0.820 & 0.804 & 0.590 \\
    \bottomrule
  \end{tabular}}
  \caption{SSIM$\uparrow$}
  \end{subtable}
  \begin{subtable}[t]{\textwidth}
  \centering
  \resizebox{0.9\columnwidth}{!}{
  \begin{tabular}{lcccccccc}
    \toprule
    Models &Trex  & Fern  & Flower  & Leaves  & Room  & Fortress  & Horns  & Orchids   \\
    \midrule
pixelNeRF & 0.618 & 0.645 & 0.658 & 0.668 & 0.603 & 0.582 & 0.669 & 0.738 \\
pixelNeRF-ft & 0.5496 & 0.6075 & 0.6335 & 0.7234 & 0.4288 & 0.5376 & 0.642 & 0.7028 \\
SRT & 0.3754 & 0.4158 & 0.4337 & 0.5559 & 0.2397 & 0.3417 & 0.4471 & 0.5375 \\
LEAP & 0.6508 & 0.7514 & 0.75 & 0.7542 & 0.5786 & 0.7135 & 0.673 & 0.7877 \\
 \midrule
PF-GRT & 0.181 & 0.208 & 0.158 & 0.285 & 0.133 & 0.136 & 0.171 & 0.312 \\
    \bottomrule
  \end{tabular}}
  \caption{LPIPS$\downarrow$}
  \end{subtable}
\end{table*}

\begin{table*}
  \caption{Comparison of PF-GRT with other pose-free generalizable novel view-synthesis methods on the Shiny datasets (scene-wise).}
  \label{tab:shiny_breakdown}
  \begin{subtable}[t]{\textwidth}
  \centering
  \resizebox{0.9\columnwidth}{!}{
  \begin{tabular}{lccccccc}
    \toprule
    Models & CD & Giants & Lab & Seasoning & Pasta & Crest & Food    \\
    \midrule
 pixelNeRF & 11.2911 & 8.7536 & 10.1085 & 9.0397 & 8.7596 & 8.6769 & 9.3571 \\
 pixelNeRF-ft & 12.5323  & 9.7544   & 11.0138  & 9.0327  & 9.6698   & 9.3726   & 10.3788  \\
 SRT & 15.5747 & 16.2062 & 11.7957 & 15.668 & 13.5123 & 12.6199 & 14.4884 \\
LEAP & 10.4334   & 8.919    & 9.4274  & 6.9765   & 10.9673  & 9.156  & 10.15   \\
     \midrule
 PF-GRT & 23.3704 & 22.1177 & 13.561 & 22.8052 & 19.1302 & 14.0699 & 19.1466 \\
 
    \bottomrule
  \end{tabular}}
  \caption{PSNR$\uparrow$}
  \end{subtable}
  \begin{subtable}[t]{\textwidth}
  \centering
  \resizebox{0.9\columnwidth}{!}{
  \begin{tabular}{lccccccc}
    \toprule
    Models &  CD & Giants & Lab & Seasoning & Pasta & Crest & Food   \\
    \midrule
pixelNeRF & 0.351 & 0.326 & 0.24 & 0.127 & 0.492 & 0.418 & 0.275  \\
pixelNeRF-ft & 0.3762 & 0.3639 & 0.3551 & 0.1757 & 0.4983 & 0.5289 & 0.3719 \\
SRT & 0.8166 & 0.7976 & 0.8637 & 0.801 & 0.7821 & 0.8849 & 0.6413  \\
LEAP & 0.2596 & 0.3175 & 0.3435 & 0.4334 & 0.1388 & 0.2476 & 0.352 \\
\midrule
PF-GTR & 0.798 & 0.737 & 0.773 & 0.674 & 0.848 & 0.820 & 0.804 \\
    \bottomrule
  \end{tabular}}
  \caption{SSIM$\uparrow$}
  \end{subtable}
  \begin{subtable}[t]{\textwidth}
  \centering
  \resizebox{0.9\columnwidth}{!}{
  \begin{tabular}{lccccccc}
    \toprule
    Models & CD & Giants & Lab & Seasoning & Pasta & Crest & Food    \\
    \midrule
pixelNeRF & 0.6849 & 0.5693 & 0.6756 & 0.6279 & 0.6499 & 0.8021 & 0.6388 \\
pixelNeRF-ft & 0.6628 & 0.5492 & 0.6704 & 0.6265 & 0.631 & 0.7897 & 0.6379 \\
SRT & 0.3308 & 0.2361 & 0.5706 & 0.3832 & 0.2284 & 0.5667 & 0.3395 \\
LEAP & 0.6449 & 0.7132 & 0.6774 & 0.6807 & 0.702 & 0.7294 & 0.6782 \\
\midrule
PF-GRT & 0.1896 & 0.2178 & 0.5078 & 0.2149 & 0.187 & 0.5232 & 0.2912 \\
    \bottomrule
  \end{tabular}}
  \caption{LPIPS$\downarrow$}
  \end{subtable}
\end{table*}

\begin{table*}
  \caption{Comparison of PF-GRT with other pose-free generalizable novel view-synthesis methods on the real Real-Estate datasets (scene-wise).}
  \label{tab:real-estate_breakdown}
  \begin{subtable}[t]{\textwidth}
  \centering
  \resizebox{0.5\columnwidth}{!}{
  \begin{tabular}{lcccc}
    \toprule
    Models & 0bcef  & 000db  & 000eb  & 8516c     \\
    \midrule
 pixelNeRF & 8.541 & 9.284 & 10.084 & 8.055 \\
pixelNeRF-ft & 11.5395 & 11.4856 & 10.7908 & 10.5445 \\
 SRT & 17.1401 & 17.3898 & 16.261 & 16.6377 \\
LEAP & 11.6061 & 12.329 & 11.3418 & 11.2685 \\
   \midrule
PF-GRT & 24.760 & 22.808 & 23.487 & 25.778 \\
    \bottomrule
  \end{tabular}}
  \caption{PSNR$\uparrow$}
  \end{subtable}
  \begin{subtable}[t]{\textwidth}
  \centering
  \resizebox{0.5\columnwidth}{!}{
  \begin{tabular}{lcccc}
    \toprule
    Models & 0bcef  & 000db  & 000eb  & 8516c     \\
    \midrule
pixelNeRF & 0.427 & 0.380 & 0.401 & 0.373 \\
pixelNeRF-ft & 0.5093 & 0.4646 & 0.48 & 0.4381 \\
SRT & 0.6594 & 0.5449 & 0.5429 & 0.6012 \\
LEAP & 0.528 & 0.5261 & 0.5256 & 0.5291 \\
\midrule
PF-GRT & 0.804 & 0.750 & 0.785 & 0.816 \\
    \bottomrule
  \end{tabular}}
  \caption{SSIM$\uparrow$}
  \end{subtable}
  \begin{subtable}[t]{\textwidth}
  \centering
  \resizebox{0.5\columnwidth}{!}{
  \begin{tabular}{lcccc}
    \toprule
    Models & 0bcef  & 000db  & 000eb  & 8516c     \\
    \midrule
pixelNeRF & 0.507 & 0.515 & 0.486 & 0.504 \\
pixelNeRF-ft & 0.4958 & 0.4694 & 0.4518 & 0.5018 \\
SRT & 0.3152 & 0.2922 & 0.3252 & 0.3134 \\
LEAP & 0.4608 & 0.4563 & 0.4408 & 0.4581 \\
  \midrule
PF-GRT & 0.174 & 0.220 & 0.193 & 0.172 \\
    \bottomrule
  \end{tabular}}
  \caption{LPIPS$\downarrow$}
  \end{subtable}
\end{table*}

\begin{table*}
  \caption{Comparison of PF-GRT with other pose-free generalizable novel view-synthesis methods on the NeRF Synthetic Datasets (scene-wise).}
  \label{tab:blender_breakdown}
  \begin{subtable}[t]{\textwidth}
  \centering
  \resizebox{0.9\columnwidth}{!}{
  \begin{tabular}{lcccccccc}
    \toprule
    Models & Chair & Drums & Ficus & Hotdog & Materials & Mic & Ship & Lego  \\
    \midrule
 pixelNeRF & 7.2024 & 7.7479 & 7.4265 & 6.9255 & 7.0926 & 7.0269 & 6.3125 & 7.1134 \\
 pixelNeRF-ft & 7.8914 & 8.3051 & 8.1891 & 7.6405 & 8.2315 & 7.591 & 7.2083 & 8.0065 \\
  SRT & 16.0348 & 15.6772 & 15.0571 & 15.8147 & 15.1039 & 14.5086 & 13.7598 & 14.417 \\
 LEAP & 17.466 & 15.2234 & 19.4337 & 17.0554 & 17.0797 & 19.4747 & 21.6511 & 16.7814 \\
  \midrule
 PF-GRT & 25.104 & 19.192 & 21.785 & 22.712 & 27.359 & 25.14 & 16.533 & 21.019 \\
 
    \bottomrule
  \end{tabular}}
  \caption{PSNR$\uparrow$}
  \end{subtable}
  \begin{subtable}[t]{\textwidth}
  \centering
  \resizebox{0.9\columnwidth}{!}{
  \begin{tabular}{lcccccccc}
    \toprule
    Models& Chair & Drums & Ficus & Hotdog & Materials & Mic & Ship  & Lego  \\
    \midrule
    pixelNeRF & 0.6046 & 0.5743 & 0.6283 & 0.6036 & 0.5708 & 0.6191 & 0.4011 & 0.5232 \\
    pixelNeRF-ft & 0.6435 & 0.6334 & 0.68 & 0.6643 & 0.6083 & 0.6564 & 0.4278 & 0.5535 \\
    SRT & 0.8166 & 0.7976 & 0.8637 & 0.801 & 0.7821 & 0.8849 & 0.6413 & 0.7349 \\
    LEAP & 0.8696 & 0.7965 & 0.9094 & 0.831 & 0.8049 & 0.9089 & 0.7531 & 0.7598 \\
     \midrule
 PF-GRT &  0.871  & 0.835  & 0.822  &0.875  & 0.8  & 0.881  & 0.677  &0.817 \\
    \bottomrule
  \end{tabular}}
  \caption{SSIM$\uparrow$}
  \end{subtable}
  \begin{subtable}[t]{\textwidth}
  \centering
  \resizebox{0.9\columnwidth}{!}{
  \begin{tabular}{lcccccccc}
    \toprule
    Models & Chair & Drums & Ficus & Hotdog & Materials & Mic & Ship & Lego \\
    \midrule
pixelNeRF & 0.3755 & 0.4147 & 0.3515 & 0.4186 & 0.4162 & 0.372 & 0.5294 & 0.4321 \\

pixelNeRF-ft & 0.3651 & 0.4076 & 0.3223 & 0.3715 & 0.3819 & 0.3453 & 0.5064 & 0.4018 \\
SRT & 0.2024 & 0.2121 & 0.1883 & 0.2318 & 0.2148 & 0.1942 & 0.2787 & 0.2505 \\
 LEAP & 0.1666 & 0.2132 & 0.1184 & 0.2169 & 0.1896 & 0.1111 & 0.246 & 0.2243 \\
  \midrule
PF-GRT & 0.083 & 0.154 & 0.098 & 0.087 & 0.174 & 0.046 & 0.402 & 0.126 \\
   
    \bottomrule
  \end{tabular}}
  \caption{LPIPS$\downarrow$}
  \end{subtable}
\end{table*}

\begin{figure*}[t]
\begin{center}
    \centering
    \includegraphics[width=0.99\linewidth]{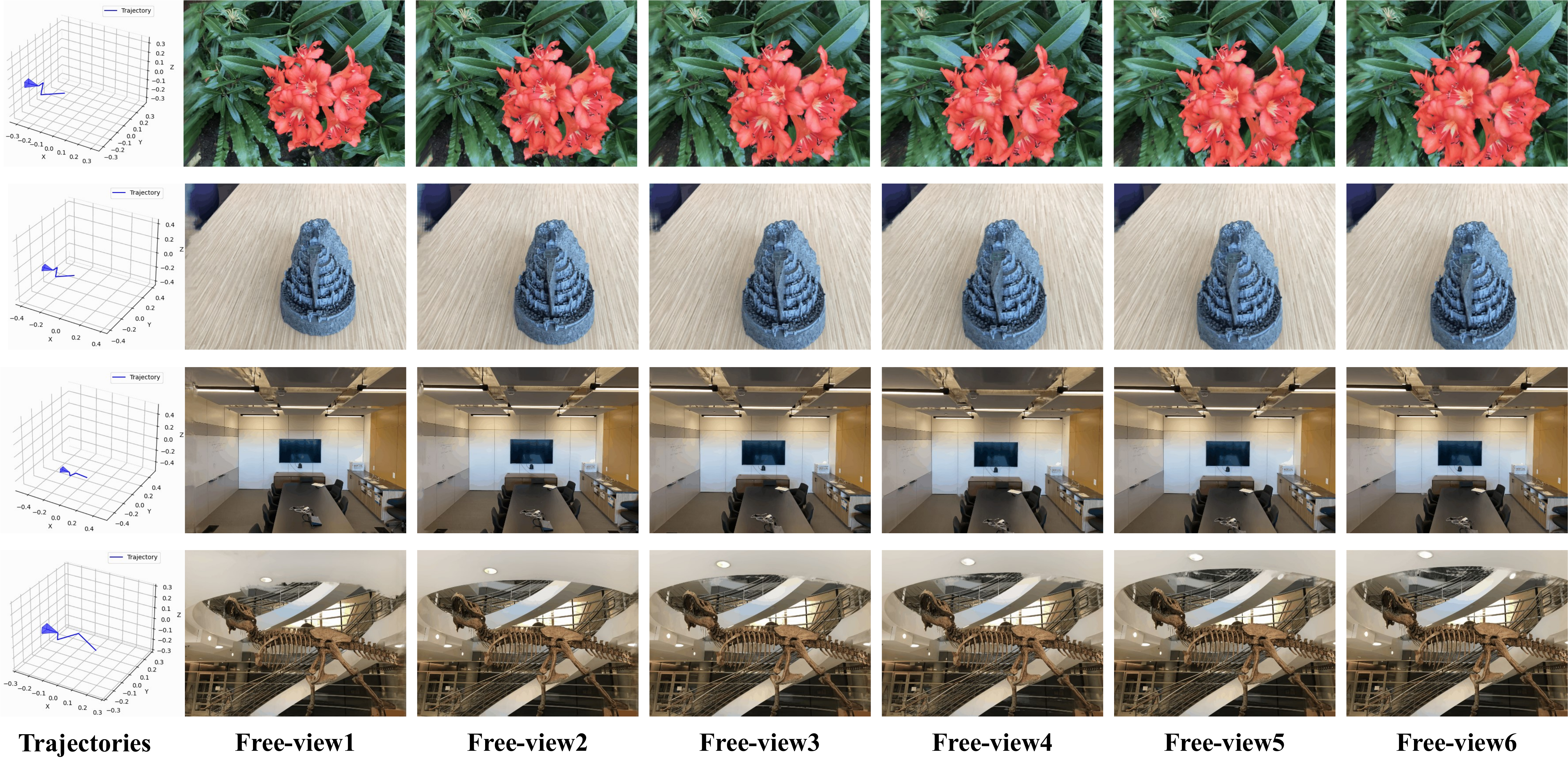} 
\caption{Visualization on more interpolated viewpoints. The visualized images showcase the efficacy of our method in handling various viewpoints by interpolating between test views on real-world datasets. See the video in supplementary materials for more detailed comparisons.}
\label{fig:free-view}
\end{center}
\end{figure*}

\paragraph{More Free-View Rendering}
We provide the visualization of multiple rendered novel viewpoints for the real-world dataset by interpolating between test views. This visualization, illustrated in Figure~\ref{fig:free-view}, demonstrates the capabilities of our method in generating diverse viewpoints, offering insight into its performance and potential limitations in real-world scenarios. 

\section{Conclusion and Limitations}

We present a novel framework, PF-GRT, for photo-realistic rendering from a sparse set of unposed images.
PF-GRT constructs a relative coordinate system to parameterize the target view.
It adapts the \textit{OmniView Transformer} for a pose-free setting, effectively fusing the unposed source images, and aggregating multi-view cues toward the origin view via a data-driven attention mechanism.
PF-GRT enjoys the advantages of global feature matching, and Image-Based Rendering (IBR) to produce state-of-the-art rendering quality in complex scenes.
Moreover, PF-GRT can render new views on unseen datasets without any scene-specific optimization and pre-computed camera poses, showcasing both the flexibility in pose annotations and robustness against noisy computed camera poses.
Our approach also carries certain limitations inherited from these previously proposed methods (e.g., IBRNet~\citep{wang2021ibrnet}). For instance, while our method requires casting rays for rendering a pixel, this inevitably introduces a computational rendering cost, similar to other MLP-based~\cite{mildenhall2021nerf, barron2021mip,Barron2022MipNeRF3U} and Transformer-based~\cite{wang2022attention,wang2021ibrnet,suhail2022generalizable} neural rendering frameworks. Incorporating depth priors from the predictive multi-view depth to guide the point sampling~\cite{lin2022efficient}, or leveraging the Gaussian Splatting~\cite{kerbl20233d} technique may mitigate this efficiency issue.



\clearpage

{
    \small
    \bibliographystyle{ieeenat_fullname}
    \bibliography{main}
}

\end{document}